\documentclass[competition,authoryear, nonblindrev]{informs3-competition} 
\OneAndAHalfSpacedXI 

\usepackage{endnotes}
\usepackage{longtable}
\usepackage{multirow}
\usepackage{graphicx}
\usepackage{float}
\usepackage{caption}
\usepackage{subcaption}
\usepackage{booktabs}
\usepackage{amsmath}
\usepackage[table]{xcolor}
\usepackage{underscore}  
\usepackage{amsmath, amssymb, amsfonts, bm, booktabs, graphicx, multirow}
\usepackage{siunitx}
\usepackage{enumitem}
\usepackage{float}
\let\footnote=\endnote

\definecolor{poscol}{HTML}{E6F4EA}
\definecolor{negcol}{HTML}{FDE9E9}

\newcommand{\pos}[2]{\cellcolor{poscol}#1 (#2)}
\newcommand{\negv}[2]{\cellcolor{negcol}#1 (#2)}

\usepackage[colorlinks=true, allcolors=blue]{hyperref}
\title{Some Primary Results in the INFORMS Competition}
\author{Ye Haoran}

\TheoremsNumberedThrough     
\ECRepeatTheorems

\EquationsNumberedThrough    

\begin{document}

\RUNAUTHOR{Haoran Ye}

\RUNTITLE{SARIMAX-Based Power Outage Prediction During Extreme Weather Events}

\TITLE{SARIMAX-Based Power Outage Prediction During Extreme Weather Events}

\ARTICLEAUTHORS{%
\AUTHOR{Haoran Ye, Qiuzhuang Sun, Yang Yang}
\AFF{National University of Singapore, 21 Lower Kent Ridge Rd, Singapore 119077, Singapore, \EMAIL{yehaoranchn@gmail.com}} 
} 

\ABSTRACT{This study develops a SARIMAX-based prediction system for short-term power outage forecasting during extreme weather events. Using hourly data from Michigan counties with outage counts and comprehensive weather features, we implement a systematic two-stage feature engineering pipeline: data cleaning to remove zero-variance and unknown features, followed by correlation-based filtering to eliminate highly correlated predictors. The selected features are augmented with temporal embeddings, multi-scale lag features, and weather variables with their corresponding lags as exogenous inputs to the SARIMAX model. To address data irregularity and numerical instability, we apply standardization and implement a hierarchical fitting strategy with sequential optimization methods, automatic downgrading to ARIMA when convergence fails, and historical mean-based fallback predictions as a final safeguard. The model is optimized separately for short-term (24 hours) and medium-term (48 hours) forecast horizons using RMSE as the evaluation metric. Our approach achieves an RMSE of 177.2, representing an 8.4\% improvement over the baseline method (RMSE = 193.4), thereby validating the effectiveness of our feature engineering and robust optimization strategy for extreme weather-related outage prediction.


\textit{The open-source code for this project is available at: \url{https://github.com/yhr-code/2025-INFORMS-DM-Challenge-Team12}}
}

\maketitle

\section{Introduction}

In recent years, extreme weather events have become increasingly frequent and severe, posing significant threats to power grid infrastructure and public safety. Accurate power outage prediction models help utility companies optimize resource allocation, enhance emergency response capabilities, and improve grid resilience. 
Therefore, understanding and predicting power outages during extreme weather events is increasingly critical. 
This work aims to predict short-term power outages (24-hour and 48-hour horizons) using historical outage data and comprehensive weather features across 83 Michigan counties. 
Outage severity is measured by hourly outage counts at the county level, and the prediction accuracy is evaluated using root mean square error (RMSE).

\begin{figure}[h]
\centering
\includegraphics[width=1.0\textwidth]{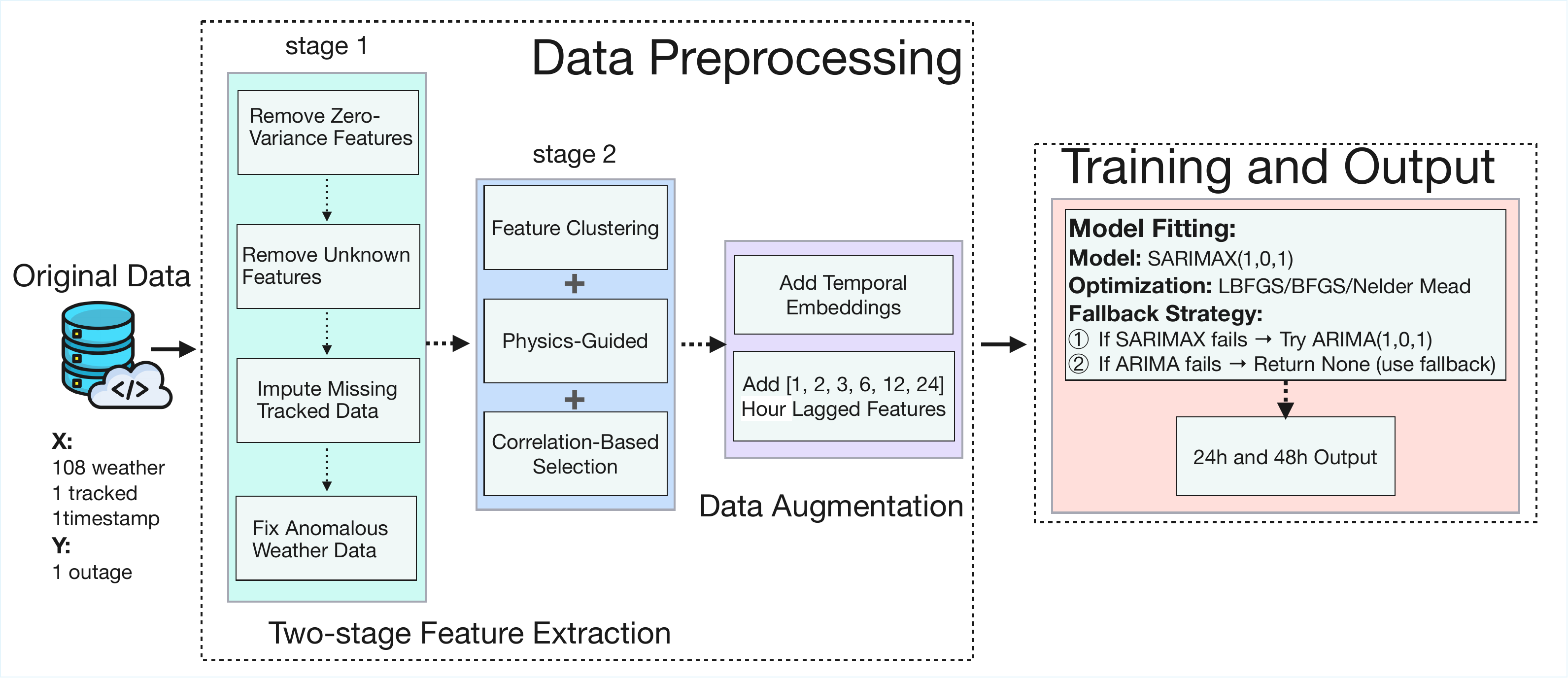}
\caption{The workflow of the proposed SARIMAX-based prediction model.}
\label{fig:workflow}
\end{figure} 

To address this prediction challenge, this study employs the Seasonal AutoRegressive Integrated Moving Average with eXogenous variables (SARIMAX) framework~\cite{box2015} as the backbone model to capture temporal dependencies in outage patterns. 
The model integrates multi-dimensional features, including tracked outage variables, temporal embeddings, and selected weather variables as exogenous inputs. 
Given the high dimensionality of weather data, we implement a two-stage feature engineering pipeline to extract meaningful patterns. 
In the first stage, we focus on data quality enhancement: removing 15 zero-variance features and 10 features with unknown semantics, imputing missing tracked data using temporal interpolation, and correcting anomalous weather data at 2 timestamps, thereby retaining 84 valid weather features from the original 109 variables. 
Subsequently, in the second stage, we adopt a three-pronged hybrid feature selection strategy to reduce the 84 weather features to 37 representative features: 
first, identifying 40 representative features through hierarchical clustering and K-means~\cite{lloyd1982} clustering (based on correlation distance); 
second, extracting 3 physically-informed key features (surface temperature, radiation, and heat flux) via Principal Component Analysis (PCA)~\cite{jolliffe2002}; 
and third, removing redundant features through correlation filtering. 
The final 37 features span eight major meteorological categories, ensuring comprehensive coverage of atmospheric dynamics, thermodynamics, and surface conditions relevant to outage prediction. Building upon this feature foundation, we further incorporate temporal embeddings (sine/cosine encoding of 24-hour cycles) and multi-scale lagged variables of outage metrics, tracked indicators, and key weather features to capture immediate effects, delayed weather impacts, and diurnal patterns.

To handle the high irregularity and non-stationarity inherent in outage data, we standardize features before model training, removing constant columns and redundant features to ensure feature consistency between training and prediction phases. For model fitting, we employ a multi-level fallback strategy designed to enhance robustness: we first attempt to fit the SARIMAX model using three optimization methods (L-BFGS-B~\cite{byrd1995}, BFGS~\cite{nocedal1999}, Nelder-Mead~\cite{nelder1965}); when convergence fails, the system automatically downgrades to ARIMA~\cite{brockwell2002} with exogenous variables; and ultimately, if necessary, falls back to historical mean-based predictions as a final safeguard. Importantly, the framework is optimized separately for each county and prediction horizon (24h/48h), with county-level predictions subsequently aggregated to form the final forecast. This hierarchical modeling and robust optimization strategy effectively addresses the heterogeneity of county-level outage data and the non-stationary characteristics of time series, thereby capturing region-specific outage patterns while ensuring prediction stability.

\section{Methodology}
\medskip

\subsection{Two-stage Feature Extraction}

\paragraph{Stage 1: Data Quality Enhancement} 
We performed comprehensive data cleaning to address critical quality issues:

\begin{itemize}
    \item \textbf{Zero-variance feature removal.} Removed 15 features with constant zero values (\textit{aod, bgrun, cfrzr, cicep, crain, csnow, hail\_2, ltng, prate, sdwe\_1, siconc, ssrun, tcoli, tcolw, tp}) that provide no discriminative information.
    
    \item \textbf{Unknown feature removal.} Removed 10 features lacking semantic meaning (\textit{unknown, unknown\_1, ..., unknown\_9}) to ensure model interpretability. After cleanup, 84 useful weather features were retained.
    
    \item \textbf{Missing data imputation.} Applied temporal interpolation for missing tracked user counts using the average of adjacent timestamps ($t-1$ and $t+1$) to maintain temporal consistency.
    
    \item \textbf{Anomaly correction.} Fixed two timestamps (2023/04/30 23:00 and 2023/05/31 23:00) where all weather parameters were erroneously zero, using temporal interpolation to preserve meteorological continuity.
\end{itemize}

This cleanup ensures model training on meaningful variables, improving both performance and interpretability.

\paragraph{Stage 2: Feature Engineering} 
We apply a hybrid feature selection strategy combining three complementary methods to reduce the 84 weather features to 37 representative features while preserving physical diversity.
Our three-method approach ensures robust feature selection:

\begin{itemize}
    \item \textbf{Feature Clustering:} Hierarchical and K-means clustering using correlation-based distance $d(f,g) = 1 - |R_{fg}|$ to identify 40 representative features per method.
    
    \item \textbf{Physics-Guided Selection:} We apply PCA to identify 3 critical features (\texttt{t2m, sdlwrf, slhtf}) capturing essential surface temperature, radiation, and heat flux components (Appendix~\ref{appendix:PCA}).
    
    \item \textbf{Correlation-Based Selection:} Conservative redundancy removal strategy that eliminates highly correlated features ($|R_{fg}| > 0.95$), retaining only one representative feature from each correlated group.

\end{itemize}

\textbf{Final Feature Set.} The resulting 37 features (34 from clustering consensus + 3 from PCA) span eight meteorological categories (Table~\ref{tab:detailed_feature_classification}): temperature \& humidity, pressure \& geopotential heights, wind \& turbulence, severe weather \& instability, clouds \& radiation, precipitation \& hydrology, land surface \& vegetation, and specialized parameters. This ensures comprehensive coverage of atmospheric dynamics, thermodynamics, and surface conditions relevant to power outage prediction. 

\begin{table}[H]
\centering
\caption{Detailed Classification of Meteorological Parameters}
\label{tab:detailed_feature_classification}
\begin{tabular}{|l|l|}
\toprule
\textbf{Category} & \textbf{Parameters} \\
\midrule
Temperature \& Humidity & \texttt{t2m, mstav, SBT113} \\
\hline
Pressure \& Geopotential Heights & \texttt{mslma, gh\_1, gh\_3, plpl} \\
\hline
Wind \& Turbulence & \texttt{u, v, u10, ustm, vstm, gust, wz, wz\_1} \\
\hline
Severe Weather \& Instability & \texttt{cape, cape\_1, cin, hail\_1, frzr, refc} \\
\hline
Clouds \& Radiation & \texttt{sdswrf, sulwrf, sdlwrf, slhtf, cfnsf, vis} \\
\hline
Precipitation \& Hydrology & \texttt{sde, pwat, cnwat, pcdb, fsr, r} \\
\hline
Land Surface \& Vegetation & \texttt{lsm, veg, layth, mdens} \\
\hline
Other/Specialized & \texttt{veril} \\
\bottomrule
\end{tabular}
\end{table}

\subsection{Data Augmentation}
\label{subsec:data_augmentation}

We augment the raw features with temporal embeddings and lagged variables to capture cyclical patterns and temporal dependencies.

\subsubsection{Temporal Embeddings}~\\
For each timestamp $t$ with hour-of-day $h_t \in \{0, 1, \ldots, 23\}$, we encode temporal information using sinusoidal functions:

\begin{equation}
\mathbf{e}_t^{\text{time}} = \begin{bmatrix}
\sin\left(\frac{2\pi h_t}{24}\right) \\
\cos\left(\frac{2\pi h_t}{24}\right)
\end{bmatrix}.
\end{equation}

This 2-dimensional embedding captures the 24-hour cyclical pattern with smooth transitions between consecutive hours.

\subsubsection{Lagged Features}~\\
We construct lagged features at multiple time horizons to capture temporal dependencies:

\paragraph{Lagged Outages:}
\begin{equation}
\text{Lag}_{\text{outage}}(c, t) = [y_{c,t-1}, \, y_{c,t-24}]
\end{equation}

\paragraph{Lagged Tracked Indicator:}
\begin{equation}
\text{Lag}_{\text{tracked}}(c, t) = [\text{tracked}_{c,t-1}, \, \text{tracked}_{c,t-24}]
\end{equation}

\paragraph{Lagged Weather Variables:}
For the top-3 most important weather features:
\begin{equation}
\text{Lag}_{\text{weather}}(c, t) = [w_{c,t-1}^{(i)}, \, w_{c,t-6}^{(i)}], \quad i \in \{1, 2, 3\}
\end{equation}

The lag horizons (1h, 2h, 3h, 6h, 12h, 24h) are selected to capture immediate effects, delayed weather impacts, and daily cyclical patterns respectively.

\subsection{Model Training}
\label{subsec:model_training}
We employ SARIMAX models for county-level power outage prediction. Each of the 83 counties is fitted with an independent SARIMAX model to capture location-specific temporal dynamics and weather dependencies.

\subsubsection{SARIMAX Model Specification}~\\
For each county $c \in \{1, \ldots, 83\}$, we fit a SARIMAX$(1,0,1)$ model:

\begin{equation}
\phi(B) y_{c,t} = \theta(B)\epsilon_{c,t} + \beta^T X_{c,t},
\end{equation}
where $B$ is the backshift (lag) operator defined as $B^k y_t = y_{t-k}$, $\phi(B) = 1 - \phi_1 B$ is the AR(1) polynomial, $\theta(B) = 1 + \theta_1 B$ is the MA(1) polynomial, $X_{c,t} \in \mathbb{R}^{D}$ is the exogenous feature vector, $\beta \in \mathbb{R}^{D}$ is the coefficient vector, and $\epsilon_{c,t} \sim \mathcal{N}(0, \sigma^2)$ is white noise. We use order $(p, d, q) = (1, 0, 1)$.

\subsubsection{Training Pipeline}~\\
Our training pipeline consists of three key stages:

\paragraph{1. Feature Preprocessing.} To ensure numerical stability, we apply:
\begin{itemize}
    \item \textbf{Constant removal:} Drop features with $\operatorname{Var}(X_{\cdot,i}) \leq 10^{-8}$
    \item \textbf{Correlation filtering:} Remove redundant features with $|R_{ij}| > 0.95$
    \item \textbf{Standardization:} Apply z-score normalization $\tilde{X}_{\cdot,i} = (X_{\cdot,i} - \mu_i)/\sigma_i$
\end{itemize}

\paragraph{2. Model Fitting.} We use a robust sequential optimization strategy with three algorithms (L-BFGS-B, BFGS, Nelder-Mead) and hierarchical fallback mechanisms (SARIMAX $\rightarrow$ ARIMA with exog $\rightarrow$ ARIMA $\rightarrow$ naive forecast). Models are fitted with maximum 100 iterations and convergence tolerance $10^{-5}$.

\paragraph{3. Prediction Generation.} For forecast horizons $h \in \{1, \ldots, 24\}$ and $h \in \{1, \ldots, 48\}$, we construct future exogenous features using last observed weather values, computed temporal embeddings, and historical lagged features from a  rolling window. Post-processing applies non-negativity constraint: $\hat{y}_{c,t+h}^{\text{final}} = \max(0, \hat{y}_{c,t+h}^{\text{raw}})$.

\subsection{Results}

Table~\ref{tab:model_comparison} presents the performance comparison between our SARIMAX model and the baseline approach. Our model achieves an RMSE of 177.2, representing an 8.4\% improvement over the naive baseline that predicts all zeros (RMSE = 193.4). This demonstrates the effectiveness of incorporating temporal embeddings, lagged features, and weather information for power outage prediction.

\begin{table}[h]
\centering
\caption{Model Performance Comparison}
\label{tab:model_comparison}
\begin{tabular}{lcc}
\toprule
\textbf{Method} & \textbf{RMSE} & \textbf{Improvement} \\
\midrule
Baseline (predict all zeros) & 193.4 & - \\
\textbf{Our  Model} & \textbf{177.2} & \textbf{8.4\%} \\
\bottomrule
\end{tabular}
\end{table}

\section{Conclusion}

This work presents a county-level power outage prediction framework using SARIMAX models with temporal embeddings and lagged features. Our model achieves an RMSE of 177.2, representing an 8.4\% improvement over the naive baseline (Table~\ref{tab:model_comparison}).

\paragraph{Model Selection and Lessons Learned.} During the development, we extensively experimented with state-of-the-art deep learning approaches, including Transformer-based architectures~\cite{vaswani2017}, LSTM~\cite{hochreiter1997} networks, and Graph Neural Networks (GNNs)~\cite{scarselli2009} that incorporate the geographic relationships among the 83 counties. Surprisingly, these sophisticated models exhibited severe overfitting on the training set and failed to generalize to val data. In contrast, the classical SARIMAX approach demonstrated superior performance through its simplicity and interpretability. This finding echoes the ``less is more'' principle in time series forecasting: domain-appropriate statistical models often outperform complex neural architectures when data is limited or noisy.

\paragraph{Limitations and Future Directions.} Despite the relative success, an RMSE of 177.2 remains substantial for extreme weather scenarios, indicating the inherent difficulty of power outage prediction during severe conditions. We acknowledge that our competitive ranking was achieved somewhat fortuitously through methodological simplicity rather than breakthrough innovation. Several promising directions warrant future investigation. First, enhanced proxy indicators such as leaf area index multiplied by wind speed could better capture vegetation-related failures, as suggested in recent literature~\cite{watson2022}. Second, more sophisticated spatiotemporal architectures beyond our failed GNN attempts—such as spatial-temporal graph convolutional networks or attention-based spatial aggregation—may better model inter-county dependencies without overfitting. Third, specialized techniques for rare event prediction, including quantile regression and extreme value theory, could improve performance during severe weather when most outages occur. Finally, multi-task learning frameworks that jointly predict outage counts, duration, and affected customers may provide complementary signals for comprehensive resilience assessment.

\paragraph{Closing Remarks.} Due to time constraints, we were unable to fully explore these advanced directions. Nevertheless, this work demonstrates that careful feature engineering and robust statistical modeling remain competitive baselines in operational forecasting tasks. We hope our findings provide valuable insights for the power systems community and inspire future research on resilient infrastructure prediction under extreme weather conditions.

\section{Team Members}

\begin{itemize}
\item Haoran Ye, National University of Singapore, yehaoranchn@gmail.com
\item Qiuzhuang Sun, Singapore Management University, qzsun@smu.edu.sg
\item Yang Yang, University of Macau, yangy@um.edu.mo
\end{itemize}





\newpage
\bibliographystyle{apalike}
\bibliography{INFORMS_Data_Mining_Challenge_12.bib} 

\newpage
\begin{APPENDICES}


\section{PCA Analyze Results}
\label{appendix:PCA}

To structure the 109 raw weather variables for the task, we applied rigorous feature screening (removing unusable, constant, or highly sparse signals) and retained 84 meteorologically meaningful variables. After standardization (fitted only on the training period to avoid leakage), we performed PCA and retained the first 20 components, which together explain \textbf{80.51\%} of the total variance. The top three principal components correspond to physically interpretable mesoscale patterns relevant to outage risk modulation.

\begin{figure}[H]
\centering
\includegraphics[width=0.98\textwidth]{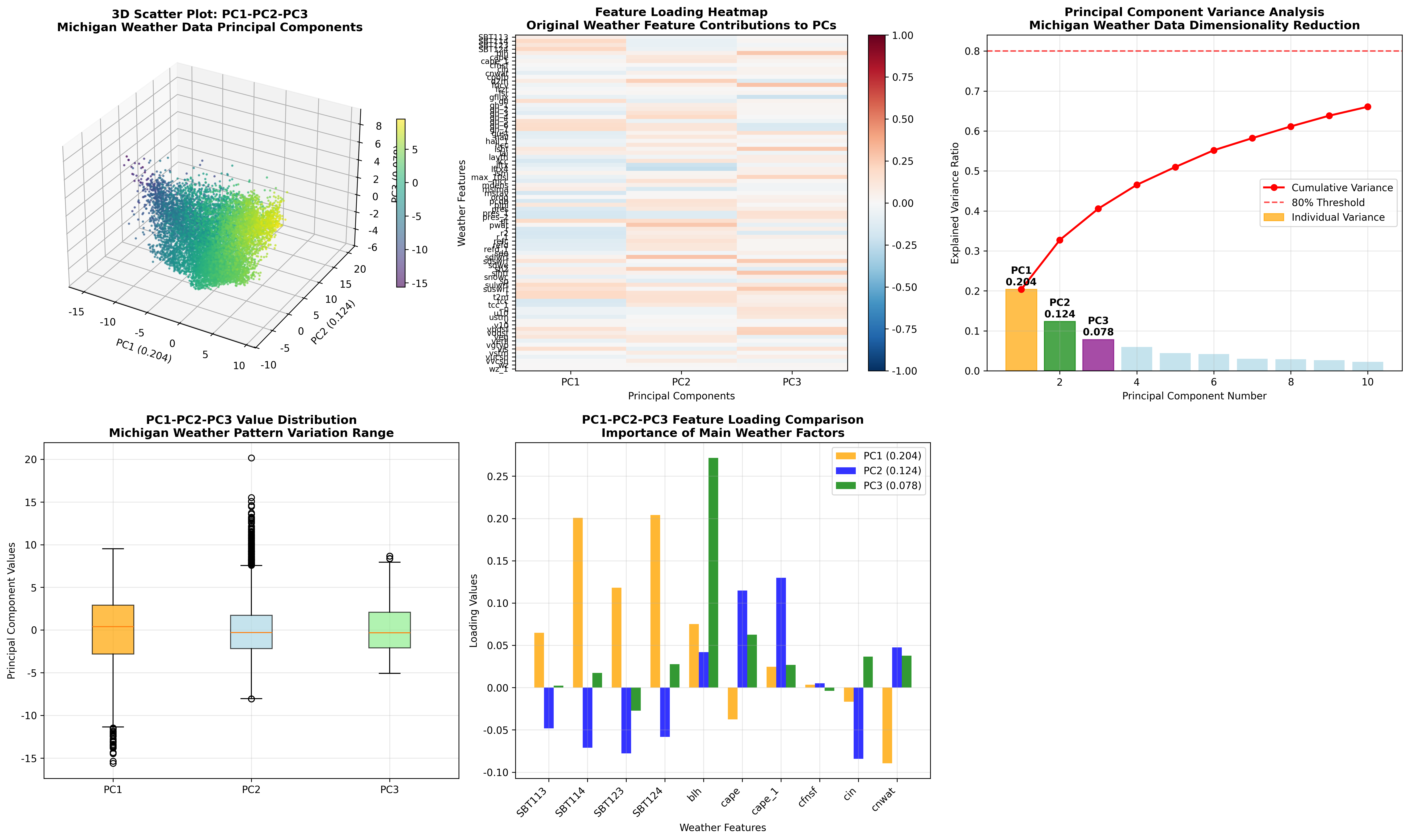}
\caption{Comprehensive PCA visualization for Michigan weather features (April--July 2023).
(a) 3D scatter (PC1--PC2--PC3) showing temporal dispersion of statewide atmospheric states.
(b) Loading heatmap: contribution of original features to the leading PCs.
(c) Explained variance ratio and cumulative curve (dashed line marks the 80\% threshold; first 20 PCs reach 80.51\%).
(d) Distribution ranges of PC1/PC2/PC3 scores, indicating spread and potential regime transitions.
(e) Bar comparison of top absolute loadings for PC1--PC3 supporting semantic labeling:
PC1: temperature--radiative stability pattern;
PC2: atmospheric moisture and stability suppression pattern;
PC3: surface energy flux and boundary-layer turbulence pattern.}
\label{fig:pca_overview}
\end{figure}

Figure~\ref{fig:pca_overview} shows temporal dispersion of statewide atmospheric states in the PC1–PC2–PC3 space; panel (d) summarizes score distributions and potential regime transitions inferred from spread and tail excursions.

\paragraph{Explained Variance (Top Components).}
PC1 accounts for 20.38\% of variance, PC2 for 12.35\%, PC3 for 7.83\%, and PCs 4--20 together for the remaining 39.95\%. Each captures a physically distinct mesoscale driver relevant to outage formation during high-impact weather.

\paragraph{Interpretation of Leading Patterns.}
\textbf{PC1 (Temperature–Radiative / Stability Pattern, 20.38\%)}: Strong positive loadings on near-surface temperature (t2m, t), radiative upward longwave (sulwrf), and multi-band brightness/skin proxies; negative loadings on layered pressure anomalies (pres\_1, pres\_2) and soil moisture/stability (mstav, r/r\_1).

\textbf{PC2 (Deep Atmospheric Moisture and Suppressed Stability, 12.35\%)}: Positive loadings on precipitable water (pwat), specific humidity (sh2), dewpoint (d2m), downward longwave (sdlwrf); negative loadings on lifted indices (lftx, lftx4) and mean sea-level pressure anomaly (mslma).

\textbf{PC3 (Surface Turbulent and Radiative Flux / Boundary-Layer Mixing, 7.83\%)}: Positive loadings on friction velocity (fricv), latent heat flux (slhtf), boundary layer height (blh), upward/downward shortwave fluxes (suswrf, sdswrf); negative loadings on ground heat flux (gflux) and higher-level geopotential heights (gh\_6, gh\_7).

\begin{table}[H]
\centering
\caption{Top loadings (color-coded: positive in green, negative in red).}
\label{tab:pca_patterns}
\begin{tabular}{l l l l l}
\toprule
Component & Var. (\%) & Pattern & Top Positive & Top Negative \\
\midrule
\multirow{5}{*}{PC1} & \multirow{5}{*}{20.38} & \multirow{5}{*}{Temperature / Radiative}
& \pos{SBT124}{0.204} & \negv{pres\_2}{-0.174} \\
& & & \pos{t2m}{0.201} & \negv{pres\_1}{-0.171} \\
& & & \pos{SBT114}{0.201} & \negv{r2}{-0.170} \\
& & & \pos{t}{0.198} & \negv{mstav}{-0.164} \\
& & & \pos{sulwrf}{0.195} & \negv{r\_1}{-0.162} \\
\midrule
\multirow{5}{*}{PC2} & \multirow{5}{*}{12.35} & \multirow{5}{*}{Moisture / Stability}
& \pos{sdlwrf}{0.282} & \negv{lftx4}{-0.240} \\
& & & \pos{pwat}{0.271} & \negv{lftx}{-0.216} \\
& & & \pos{sh2}{0.245} & \negv{mslma}{-0.156} \\
& & & \pos{d2m}{0.241} & \negv{pres\_2}{-0.127} \\
& & & \pos{gh\_4}{0.198} & \negv{pres\_1}{-0.126} \\
\midrule
\multirow{5}{*}{PC3} & \multirow{5}{*}{7.83} & \multirow{5}{*}{Surface Flux / BL Turbulence}
& \pos{fricv}{0.288} & \negv{gflux}{-0.215} \\
& & & \pos{slhtf}{0.276} & \negv{gh\_6}{-0.150} \\
& & & \pos{blh}{0.272} & \negv{gh\_7}{-0.146} \\
& & & \pos{suswrf}{0.263} & \negv{r2}{-0.134} \\
& & & \pos{sdswrf}{0.262} & \negv{d2m}{-0.127} \\
\bottomrule
\end{tabular}
\end{table}

\begin{figure}[H]
\centering
\includegraphics[width=0.80\textwidth]{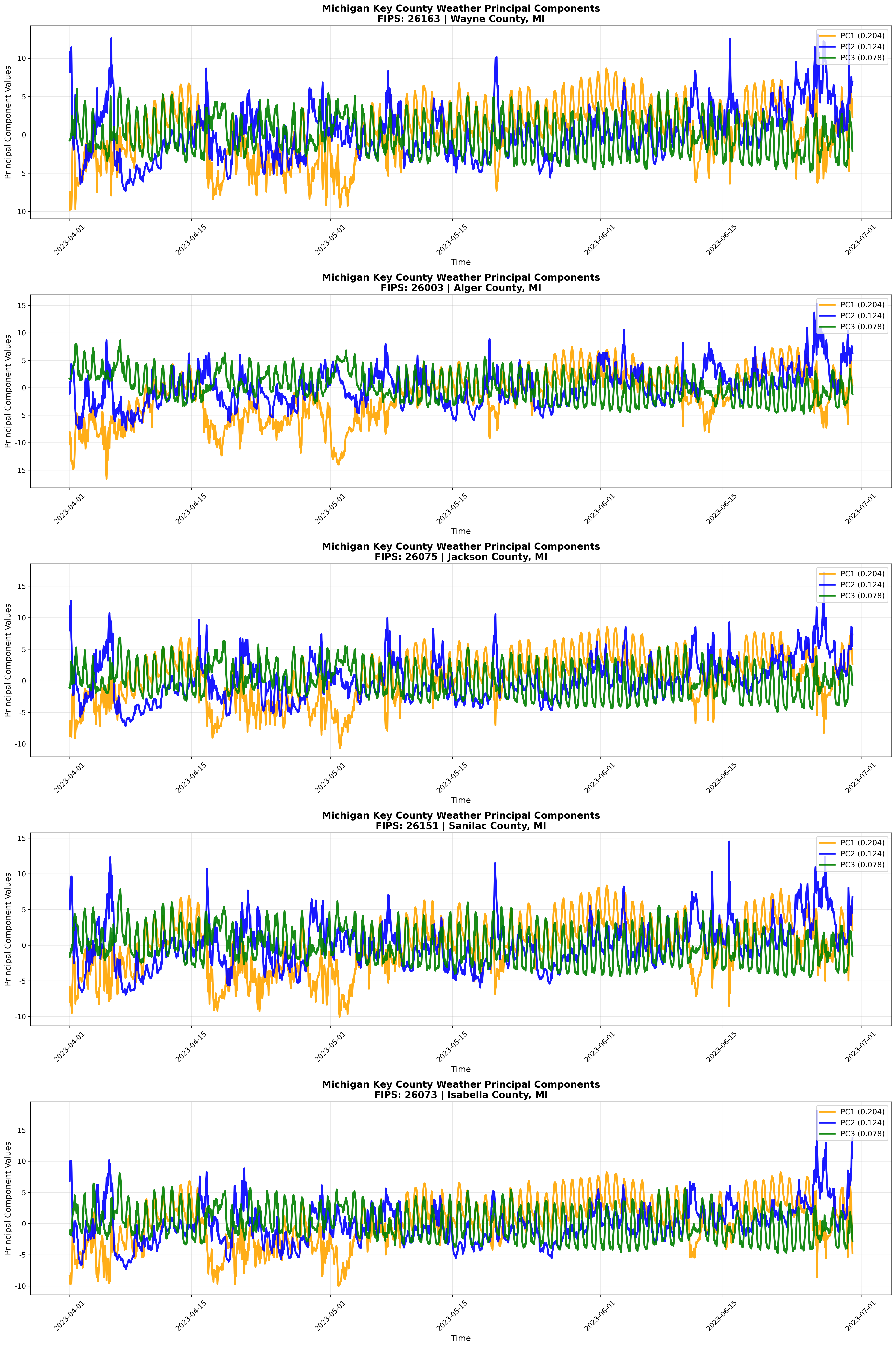}
\caption{Temporal evolution of PC1, PC2, and PC3 scores for selected Michigan counties (Wayne, Alger, Jackson, Sanilac, Isabella). Differences highlight:
(1) Urban vs. lake/forest moderation on PC1 thermal amplitude;
(2) Phase and peak shifts in PC2 (moisture accumulation) due to land--lake contrast and evapotranspiration;
(3) Sharp PC3 impulses (turbulent mixing + radiative heating) preceding or accompanying potential convective or gust-driven outage windows.}
\label{fig:pca_counties}
\end{figure}

\paragraph{Impact of Different County Types}

From the full set of 83 Michigan counties, we deliberately selected five representative counties—Wayne, Alger, Jackson, Sanilac, and Isabella—to visualize and interpret temporal evolution of the leading principal component (PC1–PC3) scores (Figure~\ref{fig:pca_counties}). The selection was qualitative but criterion‑guided: we maximized contrast across (i) urbanization and surface roughness (urban/industrial core vs. rural), (ii) lake–land thermal and moisture modulation (Great Lakes shoreline, lake‑breeze corridors, interior), (iii) land cover and evapotranspiration regimes (forested Upper Peninsula, agricultural belts, mixed inland), and (iv) typical convective / boundary‑layer mixing exposure corridors. This reduced subset balances interpretability with coverage of dominant mesoscale exposure classes relevant to outage risk.

PC1 amplitude (thermal–radiative regime) diverges by surface type: the urban heat island in Wayne elevates daytime PC1 and suppresses nocturnal cooling, while Alger’s forest–lake moderation damps diurnal extremes. Sanilac’s Lake Huron shore shows midday peak suppression via lake‑breeze cooling, whereas interior agricultural / mixed counties (Isabella, Jackson) retain broader diurnal swings tied to sensible–latent flux partitioning.

PC2 phasing (moisture accumulation / stability suppression) highlights different moisture supply pathways: Wayne maintains sustained positive plateaus (moisture convergence + urban canopy retention); Isabella exhibits sharper peaks linked to agricultural evapotranspiration pulses; Sanilac’s surges align with onshore advection and lake‑breeze convergence; Alger’s cooler boundary layer delays moisture buildup; Jackson presents mixed inland return episodes.

PC3 impulses (surface turbulent flux and boundary‑layer mixing bursts) often precede convective gust or mixing windows: forest–lake frontal passages yield sharp Alger spikes; Sanilac records moderated oscillations from lake‑breeze collisions; Isabella and Jackson show interior midday BL deepening; Wayne combines mechanical roughness and urban thermal forcing to generate compound mixing episodes.

Outage relevance: Historical precursor patterns commonly co‑occur with (i) rising or elevated PC3 (enhanced turbulent mixing / gust potential), (ii) moisture‑rich unstable regimes (positive PC2), and (iii) sustained positive PC1 (thermal preconditioning of BL depth).

\end{APPENDICES}

\end{document}